\documentclass[a4paper]{article}

\usepackage{INTERSPEECH2022}
\usepackage{multirow}
\usepackage{amsmath}
\usepackage{subcaption}

\title{Non-Linear Pairwise Language Mappings for Low-Resource Multilingual Acoustic Model Fusion}
\name{Muhammad Umar Farooq, Darshan Adiga Haniya Narayana, Thomas Hain}
\address{Speech and Hearing Research Group, University of Sheffield, UK.}
\email{mufarooq1@sheffield.ac.uk}
\begin{document}

\maketitle
\begin{abstract}

Multilingual speech recognition has drawn significant attention as an effective way to compensate data scarcity for low-resource languages. End-to-end (e2e) modelling is preferred over conventional hybrid systems, mainly because of no lexicon requirement. However, hybrid DNN-HMMs still outperform e2e models in limited data scenarios. Furthermore, the problem of manual lexicon creation has been alleviated by publicly available trained models of grapheme-to-phoneme (G2P) and text to IPA transliteration for a lot of languages. In this paper, a novel approach of hybrid DNN-HMM acoustic models fusion is proposed in a multilingual setup for the low-resource languages. Posterior distributions from different monolingual acoustic models, against a target language speech signal, are fused together. A separate regression neural network is trained for each source-target language pair to transform posteriors from source acoustic model to the target language. These networks require very limited data as compared to the ASR training. Posterior fusion yields a relative gain of 14.65\% and 6.5\% when compared with multilingual and monolingual baselines respectively. Cross-lingual model fusion shows that the comparable results can be achieved without using posteriors from the language dependent ASR.

\end{abstract}


\section{Introduction}
\label{sec:intro}

With the advancements of the computational resources, many Deep Neural Networks (DNNs) architectures and networks have been proposed to make speech recognition more efficient and accurate. DNN-HMM hybrid systems \cite{madikeri2020} outperform conventional GMM-HMM systems. For end-to-end (e2e) speech recognition, sequence-to-sequence models \cite{cho18}, RNN transducers (RNN-T) \cite{kannan2019}, transformers \cite{vishwas20} and unsupervised learning \cite{conneau21} are being used. These systems can be further improved with coupling of various techniques such as multi-task learning (MTL) \cite{sailor2020}, mixture of experts (MOE) \cite{neeraj21} and learning hidden unit contributions (LHUC) \cite{ml_epfl_thesis20} depending on the task. All these statistical modelling techniques require a lot of data for reliable parameters estimation. However, out of nearly 7000 languages being spoken around the world, just 23 languages are spoken by more than half of the world's population \cite{ethnologue}. So, sufficient data resources are available for few languages.

Over the past decade, multilingual automatic speech recognition systems have stolen the limelight being an effective way to compensate the data scarcity for low-resource languages \cite{abate20,tachbelie20,martin16,besacier14,imseng14,vu13_interspeech}. DNN based multilingual acoustic models (AM) can be used to extract features to train a monolingual model \cite{frantisek14,karel12,arnab13} or multilingual models can directly be adapted to target language \cite{tong18,huang13}. Though e2e multilingual speech recognition systems are preferred over conventional ASR to avoid lexicon creations, DNN-HMMs still outperform e2e models in limited data scenarios such as low-resource languages. Furthermore, the advancement of G2P and text to IPA transliteration approaches such as Phonetisarus \cite{phonetisarus}, Epitran \cite{epitran} and open source LanguageNet G2P models \cite{langnet} for many languages have alleviated the problem of manual creation of lexicons.

Previous work on e2e multilingual speech recognition systems has shown that a multilingual setup does not guarantee the reduction in Word Error Rate (WER) for target languages \cite{conneau21,Pratap2020,hou20}. Recent efforts to interpret the learning of multilingual speech recognition systems \cite{zelasko20,feng21} observe that Phoneme Error Rate (PER) of an overlapped phoneme is not reduced with the growing number of sharing languages. The number of shared phonemes is not a reliable metric to measure language similarities and each participating language in a multilingual system has a different similarity with the target language. Even the balanced language data sampling can cause degradation or improvement due to internal acoustic-phonetic unbalancing \cite{us}. It demands very controlled language mixing for a target language ASR.

To that end, a novel technique is proposed to fuse outputs of different monolingual models against the target language speech. Various previous studies on monolingual speech recognition have fused outputs from different models for acoustic \cite{aziz18,Ilyes17,mallidi16} and language models \cite{lmFusion1, lmFusion2}. However, monolingual models have never been fused in a multilingual setup because it can not be done straightforwardly due to different phonetic decision trees of monolingual models. In this work, a separate regression neural network (\textit{mapping network}) is trained for each \textit{$<$source, target$>$} pair to map posteriors from a source language AM to the posteriors of the target language AM. The mapped posteriors are then fused in multilingual and cross-lingual fashion for phoneme recognition of the target language. The intuition is that a \textit{mapping network} is able to learn some language related relationships between posterior distributions of source and target acoustic models. The proposed approach is helpful especially for low-resource languages because;
\begin{itemize}
    \item the \textit{mapping networks} can be trained with very limited amounts of data since a few hours can provide sufficient examples for phonetic level training.
    \item controlled fusion of posteriors based on language similarity will allow to control contribution of different source languages. 
\end{itemize}

The mapped posteriors from the monolingual AMs are fused in a multilingual setup which not only outperforms the classical multilingual systems, but also the monolingual ASRs.

\section{Acoustic Model Fusion}

\begin{figure}[tb]
\centering
\includegraphics[width=\linewidth]{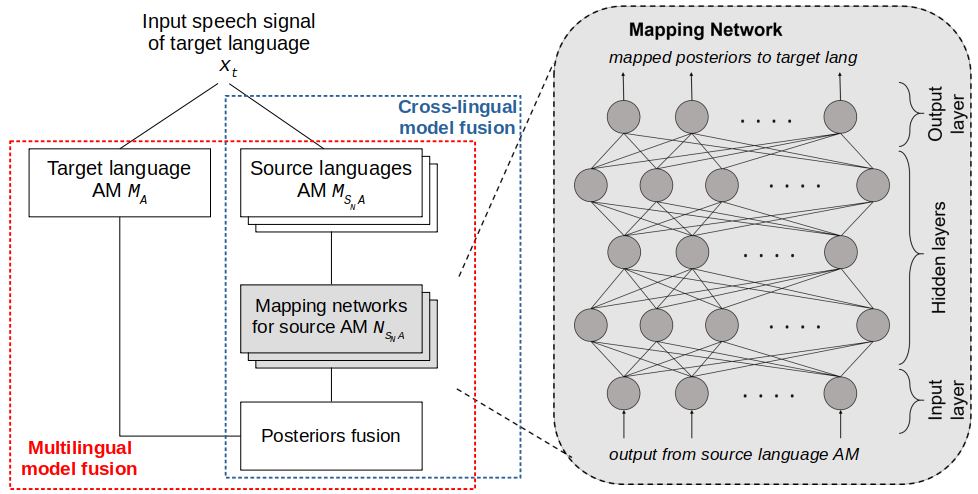}
\caption{Proposed system architecture}
\label{fig:archi}
\centering
\end{figure}

Hybrid DNN-HMM systems outperform e2e ASRs where the amount of training data is limited \cite{lfmmi}. Though the proposed fusion technique can be applied to e2e systems, the approach is described for DNN-HMM based ASRs here as a proof of concept.

In hybrid speech recognition systems, a deep neural network is trained to produce a posterior distribution of tied states of HMM models.
Theoretically, the total number of tied states  for a language with $N$ number of phonemes and $S$ number of states per HMM model is given by \(N^{n} \times S\), where $n$ is the context width. However, many polyphonemes never occur in a language and many are quite similar to the others. The total number of states is reduced by clustering many polyphonemes together. Each language yields a different phonetic decision tree in its monolingual ASR. Thus the number of tied states differs for each language and the posterior distributions are not directly comparable across the languages and thus not fusible.

Let $M_{A}$ and $M_{S_{i}}$ be the monolingual acoustic models of target and source languages respectively. A regression neural network $N_{S_{i}A}$ is trained to translate posteriors $P_{S_{i}}$ of dimension $d_{S_{i}}$ from $M_{S_{i}}$ to the posteriors $P_{S_{i}A}$ of dimension $d_{A}$ where $d_{A}$ is the dimension of posteriors from $M_{A}$. An underlying assumption is that this \textit{mapping network} is able to learn some language-related relationships between posterior distributions of source and target acoustic models. For example, the network could learn phonemes of the target language which are more amenable to cross-lingual transfer than the others.  Furthermore, a few hours of speech data can give thousands of examples that provide sufficient training data for \textit{mapping network}. The proposed system architecture is shown in Fig. \ref{fig:archi}.

Let \(X=\{x_{1},x_{2},\dotsc,x_{T}\}\) be a set of observations of target language, for which posterior distributions (\(P^{Z}=\{p_{1},p_{2},\dotsc,p_{T}\}\) where \(Z \in (A,S_{i}A) \)) are attained from all monolingual acoustic models. A mapping network is trained using KL divergence loss to map posteriors from source acoustic models ($P^{S_{i}}$) to the target language posteriors ($P^{S_{i}A}$). The loss function for a batch is given as;

\begin{equation}
\label{eq:kl}
\mathcal{L}_{S_{i}A}(\theta)=\sum_{n=1}^{N} p^{A}_{n} \cdot (\log p^{A}_{n}- \log p^{S_{i}A}_{n})
%D_{X}(M_{T},M_{S_{i}})=\frac{\sum_{t=1}^{T} p^{A}_{t} \cdot (\log p^{A}_{t}- \log p^{S_{i}A}_{t})}{T}
\end{equation}
where $N$ is the batch size for training a mapping network $N_{S_{i}A}$ to map posteriors from $i^{th}$ source language to the target language.

Posterior distributions from target AM and mapped distributions from source AMs are fused together for phoneme recognition of a target language. For a given observation at time $t$, the final posterior vector is given as;
\begin{equation}
\label{eq:wp}
p_{t}^{F}=w_{T} \cdot p_{t}^{A} + \sum_{i=1}^{K} w_{i} \cdot p_{t}^{S_{i}A}
\end{equation}
where $w_{i}$ are the scalar weights assigned to each posterior vector such that  \(\sum w_{i}=1 \) and $K$ is the number of source languages.

In the experimentation, model fusion is done in multilingual and cross-lingual settings. In cross-lingual settings, only the mapped posteriors from source models are fused (the term $w_{T} \cdot p_{t}^{A}$ is omitted from Equation \ref{eq:wp}). The cross-lingual setting avoids using target AM which is helpful for low-resource languages. The weights are assigned to the fusing languages on the basis of similarity of the source and the target language. The study on cross-lingual acoustic-phonetic similarities using the same \textit{mapping network} approach observes that the entropy of a \textit{$<$source, target$>$} mapping network shows the language similarities \cite{us}. The same similarity measure is used along with mapping network accuracy to assign the weights.

\section{Experimental Setup}
\subsection{Data set}

\begin{table}[t]
\centering
\caption{Details of BABEL data sets used for the experimentation}
\label{tab:data}
 \begin{tabular}{l|cc|cc}
\hline
\hline
\multirow{2}{4em}{Lang}&\multicolumn{2}{c|}{Train}&\multicolumn{2}{c}{Eval}\\
\cline{2-3}\cline{4-5}
&\# hours&\# spks&\# hours&\# spks\\
\hline
Tamil \textit{(tam)}&110.67&372&16.08&61\\
Telugu \textit{(tel)}&67.27&243&13.92&60\\
Cebuano \textit{(ceb)}&74.26&239&15.51&60\\
Javanese \textit{(jav)}&76.39&242&16.25&60\\
\hline
\hline
\end{tabular}
\end{table}

\begin{table}[b]
\centering
\caption{Examples (in millions) for training of mapping networks for each target language. Train set: 29 hours; Dev set: 1 hour; Eval set is same as for the ASR}
\label{tab:mndata}
 \begin{tabular}{l|ccc}
\hline
\hline
%\multirow{2}{4em}{Lang}&\multicolumn{2}{c|}{Train}&\multicolumn{2}{c}{Eval}\\
Language&Train&Dev&Eval\\
\hline
%&\# hours&\# spks&\# hours&\# spks\\
\hline
Tamil \textit{(tam)}&3.234&0.358&1.664\\
Telugu \textit{(tel)}&3.232&0.356&1.915\\
Cebuano \textit{(ceb)}&3.241&0.348&1.943\\
Javanese \textit{(jav)}&3.225&0.365&1.854\\
\hline
\hline
\end{tabular}
\end{table}

In this work, experiments are done using four low-resource languages from IARPA BABEL speech corpus \cite{babel}. Full Language Packs (FLP) of Tamil (\textit{tam}), Telugu (\textit{tel}), Cebuano (\textit{ceb}) and Javanese (\textit{jav}) are used for baseline ASR training and evaluation. Since the eval data of BABEL is not publicly available, train and dev sets of BABEL data sets are used as train and eval sets respectively for the experiments. The details of the data sets are tabulated in Table \ref{tab:data}. These data consist of conversational telephone speech and are quite challenging because of limited bandwidth, conversational styles, channel and background environment conditions. A limited amount of scripted read speech is also included in each language pack.

Full amounts are used for the training of baseline monolingual and multilingual speech recognition models. Multilingual ASR is trained by mixing data from all the languages. However, for training of the mapping networks, a subset of 30 hours is chosen from each language pack. Utterances containing only non-speech or silence are discarded while randomly sampling the 30 hours. This data is further divided randomly into 29 and 1 hour portions as train and dev sets to train the mapping networks. Since the mapping networks are trained on phonetic level, 30 hours provide millions of examples for the sufficient training of these models. The examples, used for building the mapping networks, are given in Table \ref{tab:mndata}.

\subsection{Baseline ASRs}

Baseline monolingual and multilingual acoustic models are hybrid DNN-HMM models. 40 Mel-Frequency Cepstral Coefficients (MFCCs) are extracted for each frame of the speech signals using a window size of 25ms and a shift of 10ms. These features are then fed to DNN which is consisted of 12 factorised TDNN (TDNN-F) layers \cite{ftdnn}. Each TDNN-F hidden layer is of dimension 1024, factorised with a linear \textit{`bottleneck'} dimension of 128. The acoustic model is trained using lattice-free MMI criterion (LF-MMI) \cite{lfmmi}. Neural network outputs posteriors of the clustered monophone classes. Clustering in each monolingual ASR training is different and thus the outputs from different acoustic models against an identical speech signal are not directly comparable. The experiments are done using Kaldi toolkit \cite{kaldi}.
\subsection{Mapping networks}
\begin{figure}[]
    \centering
    \includegraphics[width=\linewidth]{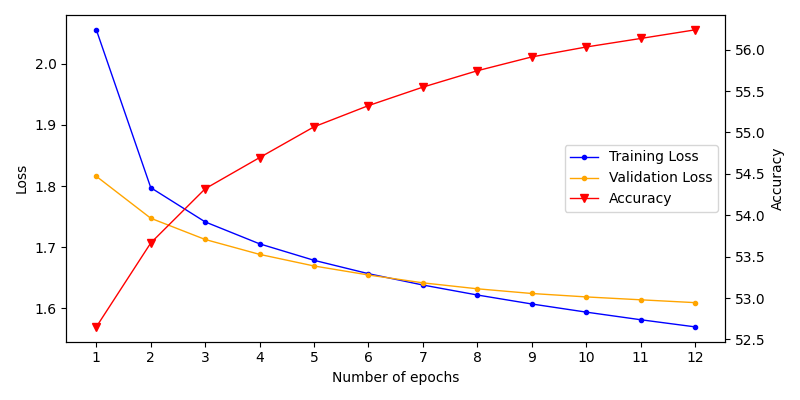}
    \caption{Training curve of $N_{ceb-tel}$ mapping network}
    \label{fig:curve}
\end{figure}

\begin{table}[b]
    \centering
    \caption{Accuracy of the mapping networks considering top n mapped classes}
    \label{tab:accu}
    \begin{tabular}{llcccc}%{p{0.1\linewidth}p{0.1\linewidth}p{0.1\linewidth}p{0.1\linewidth}p{0.1\linewidth}p{0.1\linewidth}}
    \hline \hline
        \multirow{2}{4em}{Target Lang}&\multirow{2}{4em}{Source Lang}&\multicolumn{4}{c}{\textit{Mapping network} accuracy}\\
        \cline{3-6}
        &&\textit{n=1}&\textit{n=2}&\textit{n=5}&\textit{n=10}\\
    \hline
        \multirow{3}{4em}{tam}&tel&42.91&88.16&94.91&97.91\\
        &ceb&44.43&84.63&91.82&96.13\\
        &jav&41.89&85.82&92.82&96.69\\
    \hline
        \multirow{3}{4em}{tel}&tam&54.44&92.08&96.87&98.58\\
        &ceb&35.51&90.26&95.40&97.91\\
        &jav&50.54&90.71&95.70&98.10\\
    \hline
        \multirow{3}{4em}{ceb}&tam&45.73&85.50&93.71&97.23\\
        &tel&46.17&87.87&93.98&97.51\\
        &jav&47.04&88.50&94.67&98.03\\
    \hline
        \multirow{3}{4em}{jav}&tam&47.81&85.63&93.36&96.58\\
        &tel&48.29&86.31&93.74&97.03\\
        &ceb&48.05&86.28&93.61&96.95\\
    
    \hline \hline
    \end{tabular}
    
\end{table}

\begin{figure}[t]
\centering
 \begin{subfigure}{\linewidth}
 \centering
 \includegraphics[width=\linewidth]{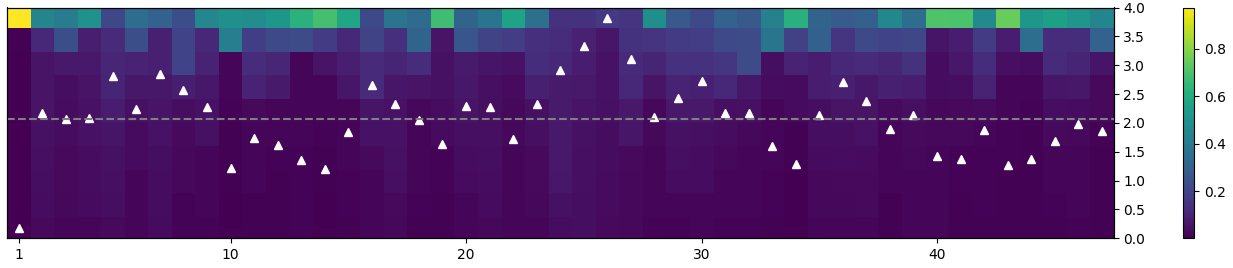}
 \caption{tam to ceb mapping network}
 %\label{fig:deen}
 \end{subfigure}
 \begin{subfigure}{\linewidth}
 \centering
 \includegraphics[width=\linewidth]{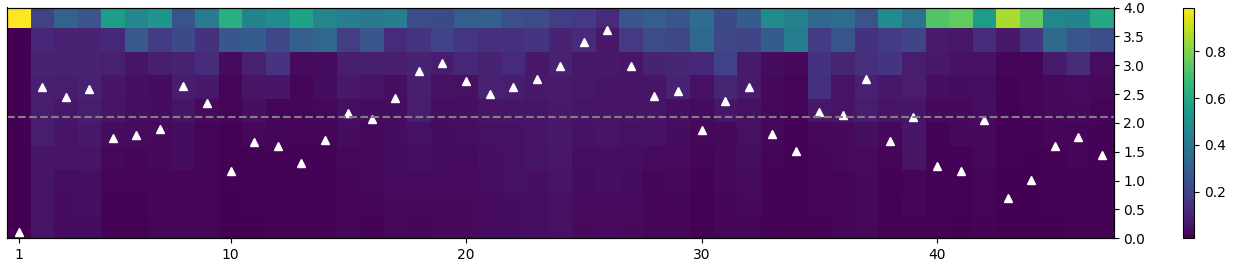}
 %\label{fig:nlen}
 \caption{tel to ceb mapping network}
 \end{subfigure}
 \begin{subfigure}{\linewidth}
 \centering
 \includegraphics[width=\linewidth]{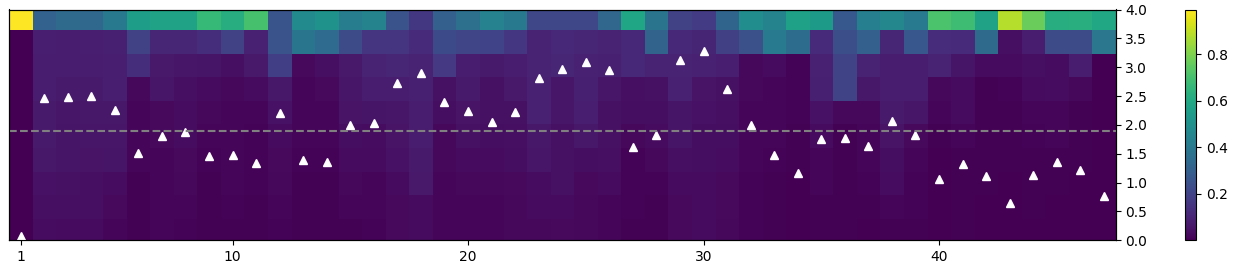}
 %\vspace{-0.5em}
 \caption{jav to ceb mapping network}
 \end{subfigure}
 \vspace{-1em}
\caption{Posteriorgram and entropy plot of $N_{tam-ceb}$, $N_{tel-ceb}$ and $N_{jav-ceb}$ for a sample $ceb$ target speech utterance. Average entropy over the utterance is plotted in grey dashed line. Each box represents a frame on horizontal axis and probability of an output class on vertical axis}
\label{fig:entropy}
\end{figure}

A regression neural network (\textit{mapping network}) is trained for each source-target language pair ($N_{src-tgt}$). The neural network consists of three fully connected hidden layers to map the posterior distributions from dimensions of $M_{S_{i}}$ to that of $M_{A}$. KL divergence loss is used for the training. As an accuracy measure of the mapping network, the number of correctly mapped frames is divided by the total number of frames as given in Equation \ref{eq:acc}. Correctly mapped frames are defined as the frames where the index of $max(mapped\_posteriors)$ is same as the index of $max(target AM\_posteriors)$ which means that the output of the mapping network is mapped to the correct clustered phone class of the target AM.
\begin{equation}
    \label{eq:acc}
    (index(max(p^{A}_{t}))==index(max(p^{S_{i}A}_{t})))\Rightarrow CMF\mathrel{+}+
\end{equation}
\begin{equation*}
    Accuracy=\frac{CMF}{T}
\end{equation*}

where $index(x)$ returns the index of class $x$ in the output vector, $CMF$ is the number of correctly mapped frames and $T$ is the total number of frames. The training curve with accuracy measure for one of the mapping networks is shown in Fig \ref{fig:curve}. Having millions of examples, training converges in early epochs for all the networks with accuracy of nearly 50\%. However, analysis reveals that most of the times when mapping network is not accurate according to the aforementioned criteria, still the correct target AM class is usually among a few most probable classes of the mapped distribution. So, the network accuracy is recalculated considering top \textit{n} classes of the \textit{mapping network} output. The results in Table \ref{tab:accu} show that the accuracy for most of the networks is dramatically increased to nearly 90\% from less than 50\% by considering top two most probable classes of the \textit{mapping network} output rather than only one.

For weighted fusion of the posteriors from different acoustic models, entropy of each \textit{$<$source, target$>$} mapping network is measured as a similarity measure. Entropy and posteriorgrams from mapping networks are shown for a sample \textit{ceb} target file in Fig \ref{fig:entropy}. Each box represents a frame on horizontal axis and probability of an output class on vertical axis. Output classes on vertical axis are sorted (from top to bottom) for each frame. Entropies of frames are plotted on right vertical axis. The figure shows that \textit{jav} to \textit{ceb} mapping network has lower average entropy than mappings from \textit{tam} and \textit{tel} AMs. This is comprehensible due to the fact that \textit{ceb} and \textit{jav} are from same language family and thus \textit{jav} is closer to \textit{ceb} than \textit{tam} and \textit{tel}. For model fusion, entropies and accuracy of the mapping networks are considered while assigning the weights. The entropies for all the trained mapping networks are tabulated in the Table \ref{tab:entropy}.

\begin{table}[t]
\centering
\caption{Entropy of the source-target mapping networks on eval set}
\label{tab:entropy}
\begin{tabular}{ccccc}
\cline{1-5}
\multirow{2}{4em}{Target Lang.}&\multicolumn{3}{c}{Source Lang.}\\
\cline{2-5}
&tam&tel&ceb&jav\\
\cline{1-5}
tam &0&1.292&1.286&1.383\\
tel &1.109&0&1.161&1.139\\
ceb &1.214&1.235&0&1.098\\
jav &1.335&1.460&1.279&0\\
\cline{1-5}
\end{tabular}
\vspace{-1em}
\end{table}

\section{Results and Discussion}
\subsection{Baseline ASRs}
Monolingual baseline systems (\textit{mono}) are the language dependent acoustic and pronunciation models which are trained on a language specific data set. The train sets of all the languages are then mixed to train the multilingual models (\textit{multi}). The results of the baseline systems are given in the Table \ref{tab:baseline} (in terms of PER). The results show that the error for all the languages is slightly increased in the baseline multilingual setup.

\subsection{Model fusion}

\begin{table}[b]
\centering
\caption{Baseline ASR performance in terms of \% PER}
\label{tab:baseline}
 \begin{tabular}{cccc|c}
\hline
\hline
Lang&\textit{mono}&\textit{multi}&\textit{multi-mf}&\textit{cross-mf}\\
\hline
tam&43.96&43.67&\textbf{41.96}&55.47 \\
tel&43.66&46.36&\textbf{42.05}&52.76 \\
ceb&36.67&41.02&\textbf{35.54}&43.04 \\
jav&41.60&45.54&\textbf{38.87}&47.79 \\
\hline
\hline
\end{tabular}
\end{table}

\begin{table}[t]
\centering
\caption{Performance of model fusion in cross-lingual setting. `Y' represents the source languages being fused together}
\label{tab:cmf}
 \begin{tabular}{p{0.2\linewidth}ccccp{0.2\linewidth}}
\hline
\hline
\multirow{2}{4em}{Target Language}&\multicolumn{4}{c}{Fused languages}&\multirow{2}{4em}{\% PER} \\
&\textit{tam}&\textit{tel}&\textit{ceb}&\textit{jav}&\\
\hline
\hline
\multirow{4}{4em}{\textit{tam}}&N&Y&Y&Y&\textbf{55.47}\\
&N&Y&N&N&55.65\\
&N&N&Y&N&57.69\\
&N&N&N&Y&57.33\\
\hline
\multirow{4}{4em}{\textit{tel}}&Y&N&Y&Y&52.76\\
&Y&N&N&N&\textbf{52.37}\\
&N&N&Y&N&55.68\\
&N&N&N&Y&53.58\\
\hline
\multirow{4}{4em}{\textit{ceb}}&Y&Y&N&Y&\textbf{43.04}\\
&Y&N&N&N&45.94\\
&N&Y&N&N&45.28\\
&N&N&N&Y&43.91\\
\hline
\multirow{4}{4em}{\textit{jav}}&Y&Y&Y&N&\textbf{47.79}\\
&Y&N&N&N&48.40\\
&N&Y&N&N&48.90\\
&N&N&Y&N&48.25\\
\hline
\hline
\end{tabular}
\end{table}

A multilingual acoustic model is imitated by fusing the target language and mapped source language posteriors. The fusion is the linear weighted sum of all these posterior distributions. In Table \ref{tab:baseline}, the results of multilingual and cross-lingual model fusion settings (\textit{multi-mf} and \textit{cross-mf} respectively) are compared with \textit{mono} and \textit{multi} baseline ASRs. Multilingual model fusion yields a maximum gain of 6.5\% over monolingual and 14.65\% when compared with multilingual baseline systems.

Results of cross-lingual model fusion shows that without using the language dependent ASR, a comparable phoneme error rate for a target language can be achieved. For cross-lingual fusion, mapped posteriors from all the source language AMs are fused. However, the computation cost for fusing large number of languages incites us to minimise the number of fusing languages. For a given target language, further experiments are carried out using the mapped posteriors from only one source language at a time. Table \ref{tab:cmf} shows that nearly similar results as \textit{cross-mf} can be achieved using mapped posteriors form the closest source language AM only. In the case of Telugu language, mapped posteriors from single AM model of Tamil perform even better than the cross-lingual model fusion. The first row for each language is same as \textit{cross-mf} of Table \ref{tab:baseline} and following rows are the cross-lingual mapped posteriors from only one of the source language AM. For a target language, the change in results using different source language AMs can be seen in relation with mapping networks entropy of Table \ref{tab:entropy}. For example in the case of Javanese target language, the entropy is highest for \textit{tel-jav} mapping network and so the WER is highest for jav when using mapped posteriors from tel AM and so on.

For the model fusion, the weights are manually assigned to the posteriors which pose an issue of sub-optimal output. However, these weights could be learnt with the training of the mapping networks. Our next steps will include expanding the work for e2e ASRs and learning the weights jointly.

\section{Conclusion}
In this work, a novel monolingual acoustic model fusion technique is proposed for low resource languages in a multilingual setup. Posterior distributions from different monolingual acoustic models against a target language speech signal are fused together. A separate regression neural network is trained for each source-target language pair to map posteriors from source acoustic model to the target language acoustic model. The mapping networks need very limited amount of data for training as compared to the ASR building. Multilingual model fusion yields a relative gain of 14.65\% and 6.5\% when compared with multilingual and monolingual baselines for the target language. Cross-lingual model fusion shows that the comparable results can be achieved without using the target language ASR.

\section{Acknowledgements}

This work was partly supported by LivePerson Inc. at the Liveperson Research Centre.

\bibliographystyle{IEEEtran}
\bibliography{template}

\end{document}